\documentclass[letterpaper, 10 pt, conference]{ieeeconf} 

\IEEEoverridecommandlockouts                              

\usepackage{graphics} 
\usepackage{epsfig} 
\usepackage{mathptmx} 
\usepackage{times} 
\usepackage{amsmath} 
\usepackage{amssymb}  
\usepackage{xcolor}
\usepackage{algorithm2e}
\makeatletter
\renewcommand{\@algocf@capt@plain}{above}
\renewcommand{\algocf@caption@plain}{\box\algocf@capbox\vskip\AlCapSkip}%
\makeatother

\usepackage[utf8]{inputenc}
\usepackage[pdfauthor={Bonetto et al.},
            pdftitle={Active  Visual  SLAM  with  Independently  Rotating  Camera},
            pdfsubject={Active V-SLAM},
            pdfkeywords={active slam,visual slam,independent camera rotation,indoor mobile robots,mapping,localization,state estimation}]{hyperref}
\let\labelindent\relax
\usepackage{enumitem}
\usepackage{subfig}
\usepackage{cuted}
\usepackage{multirow}
\hypersetup{
    colorlinks=true,
    linkcolor=blue,
    filecolor=magenta,      
    urlcolor=blue,
}

\usepackage[font=footnotesize,labelfont=bf]{caption}
\usepackage{tikz}
\def\tick{\tikz\fill[scale=0.4](0,.35) -- (.25,0) -- (1,.7) -- (.25,.15) -- cycle;} 

\usepackage[textsize=tiny]{todonotes}

\newcommand{\uproman}[1]{\uppercase\expandafter{\romannumeral#1}}

\newcommand{\fig}[1]{Fig.~\ref{#1}}
\newcommand{\tab}[1]{Tab.~\ref{#1}}

\usepackage{soul}

\title{\LARGE \bf
Active Visual SLAM with Independently Rotating Camera
}

\author{Elia Bonetto$^{1,2}$, Pascal Goldschmid$^{2}$, Michael J. Black$^1$ and Aamir Ahmad$^{2,1}$
	\thanks{$^1$Max Planck Institute for Intelligent Systems, Tübingen, Germany. {\tt\footnotesize {firstname.lastname}@tuebingen.mpg.de}}
	\thanks{$^2$Institute for Flight Mechanics and Controls, The Faculty of Aerospace Engineering and Geodesy, University of Stuttgart, Stuttgart, Germany. {\tt\footnotesize {firstname.lastname}@ifr.uni-stuttgart.de}}%
	\thanks{The authors thank the International Max Planck Research School for Intelligent Systems (IMPRS-IS) for supporting Elia Bonetto and Pascal Goldschmid.}
	\thanks{The authors would like to thank Heiko Ott and Mason Landry for their help with the design of the hardware.} %
	\thanks{978-1-6654-1213-1/21/\$31.00 \textcopyright 2021 IEEE}
}

\begin{document}

	\maketitle

\begin{abstract}
In active Visual-SLAM (V-SLAM), a robot relies on the information retrieved by its cameras to control its own movements for autonomous mapping of the environment. Cameras are usually statically linked to the robot's body, limiting the extra degrees of freedom for visual information acquisition.
\
In this work, we overcome the aforementioned problem by introducing and leveraging an independently rotating camera on the robot base. This enables us to continuously control the heading of the camera, obtaining the desired optimal orientation for active V-SLAM, without rotating the robot itself.
\
However, this additional degree of freedom introduces additional estimation uncertainties, which need to be accounted for. We do this by extending our robot's state estimate to include the camera state and jointly estimate the uncertainties.
\
We develop our method based on a state-of-the-art active V-SLAM approach for omnidirectional robots and evaluate it through rigorous simulation and real robot experiments.
\
We obtain more accurate maps, with lower energy consumption, while maintaining the benefits of the active approach with respect to the baseline.
\
We also demonstrate how our method easily generalizes to other non-omnidirectional robotic platforms, which was a limitation of the previous approach. Code and implementation details are provided as open-source.

\end{abstract}


	\section{INTRODUCTION}
\label{sec:intro}


Using RGB cameras to allow a robot to simultaneously self-localize and map an environment is a popular approach~\cite{CadenaCarloneCarrilloLatifScaramuzzaNeiraReidLeonard2016}. There exist many passive V-SLAM methods that solve this problem, e.g.~\cite{labbe2019rtab,rosinol2020kimera}, but active V-SLAM has been gaining popularity~\cite{us, rrt-uncertainty-aware, Ayoung2015} in recent years. 

In general, the cameras are rigidly attached to the robot's body constraining its \textit{relative} orientation, even if some exceptions do exist since the seminal work of~\cite{davison1998Mobile}.\footnote{Note that in hand-held mapping the camera is moving in a 3D world and represent the whole system} This design has several shortcomings. Firstly, it reduces the freedom of movement of the robotic platform. Having a fixed camera pose constrains the robot's possible movement space not only for Active Visual-SLAM approaches like~\cite{us} but also for next best view planners~\cite{rrt-uncertainty-aware}. Indeed, having a robot capable of independently controlling the camera’s orientation increases its flexibility in both mapping and control. 

Secondly, safety limits must be enforced to avoid collisions (with both environment and humans) and any possible tumble of the platform due to sudden changes in the velocities. This usually translates into low translation and rotation velocities that also reduce inertia-related movements. However, for the camera that is attached to an independent rotational joint, we can assume collision free movements and obtain a finer control due to the lighter system that needs to be rotated.

Thirdly, moving and rotating the robot implies a high energy consumption while rotating a system formed only by a camera is much more efficient.

\begin{figure}
	\centering
	\includegraphics[width=0.42\textwidth]{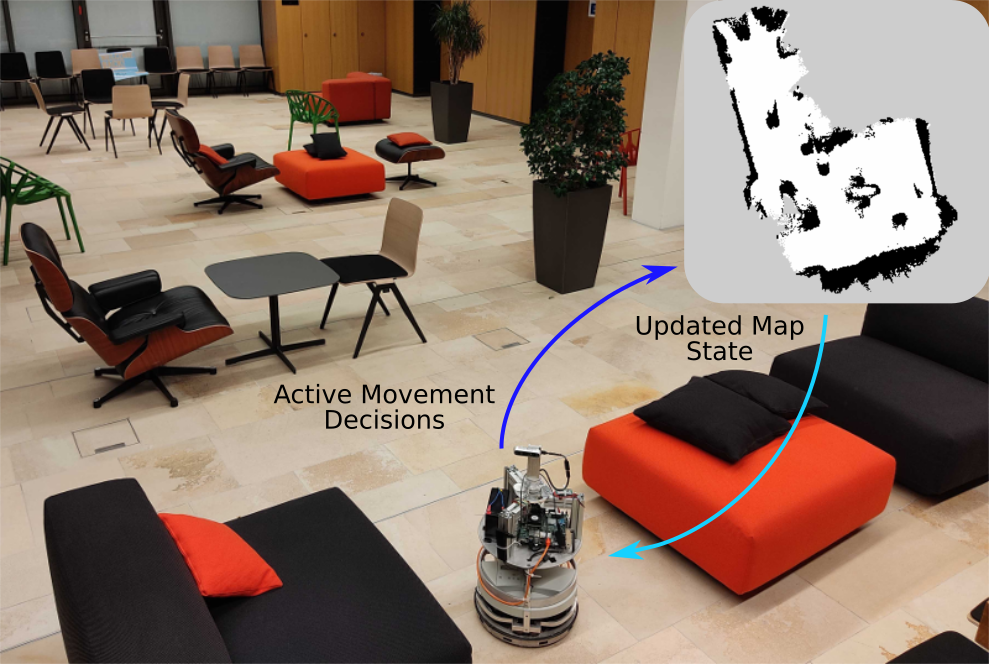}
	\caption{Our omnidirectional robot actively mapping an office reception area.}
	\label{fig:cover}
\end{figure}

Nonetheless, adding this additional degree of freedom introduces also a problem. The camera orientation w.r.t. the robot's base should be expressed through an estimate and an uncertainty. Systems that utilize these kind of rotating cameras usually disregard this problem by assuming perfect calibration and high end encoders, or by ignoring the correlation between the two systems. These assumptions are usually costly, time consuming, and not always possible nor verifiable throughout the whole experiment. We address this issue by combining the camera and the robot's pose estimates in a unified one that is then used by the system. This enables us not only to include all the information that we have with the corresponding uncertainty but also to not modify the underlying SLAM framework, thus keeping our method independent from it. 

The summary of our contributions is as follows.
\begin{itemize}
    \item We introduce a novel hardware architecture to address all the aforementioned restrictions, obtain more accurate maps and reduce energy consumption when used with~\cite{us}.
    \item We expand \textit{iRotate}~\cite{us} to different platforms.
    \item We propose a novel state estimate formulation for this kind of hardware architecture.
\end{itemize}


We test the presented method both in simulation and on a real robot. We directly compare our work to~\cite{us} by using the same simulation environments, base robot platform, and parameters. We expect similar results in the map accuracy score (since the exploration strategy is unchanged), while improvements are expected and sought in terms of entropy, trajectory error, and, more importantly, energy consumption. The real robot, a Festo Didactic's Robotino, operates in an office environment (Fig.~\ref{fig:cover} and \ref{img:robotschmema}). The movement of the camera is controlled through a single low-cost incremental encoder. Our method source code and instructions for use are available open-source\footnote{\url{https://github.com/eliabntt/active_v_slam}}.

The rest of the paper is structured as follows. In Sec.~\ref{sec:soa} we review the current state-of-the-art in independent camera movement. Our method is described thoroughly in Sec.~\ref{sec:approach}, while the experimental results are presented in Sec.~\ref{sec:exp}. In Sec.~\ref{sec:conc} we conclude with comments on future work.

	\section{Related Work}
\label{sec:soa}

Optimized cameras' orientation has been used in many scenarios. From person tracking and surveillance~\cite{1570550,6027312,carr2013hybrid}, to next best view planning and active SLAM~\cite{us,Dai2020,Schmid2020,4636757}, 3D reconstruction~\cite{pervolz2004automatic} and others. Nonetheless, despite the clear advantages that controlling the camera's direction brings, all those approaches usually lack either the independent camera orientation w.r.t. the robotic platform, or the simultaneous robot and camera's movements, or to address the uncertainty that the estimate of the camera's gaze might bring in the system.

In the simplest scenario, the camera can be considered \textit{as} the `robotic' system itself, e.g.~\cite{6027312,carr2013hybrid}, where the pan-tilt movement made by the camera is performed from a fixed global location. In those, even if the pan-tilt(-zoom) system of the camera is actively controlled, there is no other movement involved.

Other approaches, like~\cite{zdevsar2019vision}, utilize the pan-tilt unit only during the calibration of the system. While the robot is moving and localizing itself, the (camera) orientation is fixed all the time. A notable exception is~\cite{pervolz2004automatic}, where the authors propose a system of two RGB pan-tilt units. Those are used to gather only the color information for a 3D reconstruction system. In this study, cameras can reach only a set of pre-defined poses. The use of high-accuracy encoders ensures high precision repeatability of the movements lacking any form of real activeness, SLAM capabilities, or simultaneous control of the robot and camera's movement.

More advanced work, which includes a simultaneous and independent movement of the camera with respect to the robot, are presented when humanoid robots are being employed as in~\cite{10.1007}. There, the pan-tilt movement of the head, and therefore of the camera, is utilized to help the robot's localization and tracking of targets. Moreover, the link between the two is not accounted for. Here the camera, and not the whole platform, is the focus of the working system employed in a partially-known environment.

In (active) visual SLAM most approaches rely upon static cameras attached to the robotic system. Thus, even if an active heading control is employed, it is the robot that rotates and not the camera. The seminal work from Davison~\cite{davison1998Mobile} started the active feature tracking in SLAM, by using one of the first independent rotating heading mechanisms. This and the subsequent SLAM procedures developed by Davison et al. focus on the feature's tracking to reduce localization uncertainty applied to hand-held cameras. In contrast, the camera in this work is not used to track features `passively', but lies within an Active SLAM framework that seek optimal viewpoints and, therefore, is not used just to improve localization. Moreover, unlike us, they assume that the transformation matrices between the cameras and the robot are known. Fintrop et al. in~\cite{4636757} propose an actual active gaze control for SLAM using artificial landmarks, but do not introduce either a full active SLAM approach or a connection between camera and robot's odometry. This work is very similar to~\cite{manderson2016texture}, where Manderson et al. too actively control the gaze of the system while the navigation is performed by an external operator and the estimation is done for the camera frame itself. This implies that the system cannot be autonomous as it is proposed because the actual robot's orientation is unknown. Both of these methods are more related to actively control the camera rather than real active SLAM approaches. Even~\cite{4209718}, where the camera system is controlled separately w.r.t. the robot, gaze prediction is agnostic about robot's movements and uses a greedy policy in the control, therefore lacking any link between the two and almost performing the two tasks separately. In~\cite{us} Bonetto et al. leverage the omnidirectional nature of the robot. In~\cite{rrt-uncertainty-aware}, instead, it is the high mobility of the drone that is being used to exploit a continuous change of heading limiting both approaches with respect to the considered platforms. Applying~\cite{us} to a non-omnidirectional robot or~\cite{rrt-uncertainty-aware} to other aerial vehicles like blimps is not straightforward without a separate camera control. It would be possible to utilize either an array of cameras or a $360^\circ$ one. However, in this work we wanted to keep the setup as similar as possible to the baseline. Moreover, using that kind of setup, apart for being costly, usually poses a series of challenges in terms of computational requirements and processing of the data due either to the distortion introduced or the amount of data and the synchronization necessary. Ultimately, to the best of our knowledge, there is yet no method that combines a full active SLAM approach with an active camera movement that is independent of the robotic platform \textit{and} considers the system as a whole.

	\section{Proposed Approach}
\label{sec:approach}
\subsection{Problem Description and Notations}

Let there be a robot, traversing on a 2D plane in a 3D environment, whose pose is given by $\mathbf{x}_R = [x_R ~~ y_R ~~ \theta_R]^{\top}$ and velocity by $\mathbf{\dot{x}}_R = [\dot{x}_{R} ~~ \dot{y}_{R} ~~ \dot{\theta}_{R}]^{\top}$. Let this robot be equipped with an RGB-D camera with an associated maximum sensing distance $d_{thr}$ and horizontal field of view (FOV) $\alpha$, a 2D laser range finder, and two IMU units, one of them attached to the camera and one to the robot's base. The camera is mounted on a joint that allows a full $360^{\circ}$ independent rotation along the robot's z-axis, i.e. it has independent pan movement. We denote the orientation of the camera with respect to the robot as $\gamma_{C}^{R}$ and its rotational velocity as $\dot{\gamma}_C^{R}$. Therefore, the orientation of the camera with respect to the world frame is $\psi = \theta_{R} + \gamma_{C}^{R}$ and its velocity is $\dot{\psi} = \dot{\theta}_{R} + \dot{\gamma}_{C}^{R}$.

Let $M \subset \mathbb{R}^2$ then represent a bounded 2D grid map of the environment, where for each point $\mathbf{m} = [x_m ~~ y_m]^{\top} \in M$ we have an occupancy probability $p_o(\mathbf{m}) \in [0,1]$. 

We assume that the robot begins exploration from a collision-free state and all map cells belong to the \textit{unknown} space $M_{unk} \subset M$. 

The robot's goal is to autonomously map all the observable points in $M_{unk}$ as \textit{free} ($M_{free}$) or \textit{occupied} ($M_{occ}$). 

At any given instant the robot's state, map state, the set of observed visual features, and the SLAM graph $\mathbf{G}$ of previously visited locations are available to the robot. A node $\mathbf{n} = [x ~~ y ~~ \theta]^{\top} \in \mathbf{G}$ is defined by its coordinates in the map frame and by the orientation of the robot. While solving this problem, the robot must also be able to avoid both static and dynamic obstacles. The active SLAM approach used within this work is the one depicted in~\cite{us}. 

The goal is to use the camera's independent rotation to allow a simultaneous turn of both the camera and the robot to reduce the overall energy consumption thanks to the increased freedom of movement. Moreover, we seek to apply~\cite{us} to different, non-omnidirectional, robotic platforms while keeping at least comparable performance. Finally, we will study the effect that the merging of the estimation of the camera and the robot poses has, seeking better performance in terms of the final map's entropy, accuracy, and trajectory error w.r.t. the corresponding non-combined alternative. 

\subsection{Kinematic Description}
\label{subsec:KIN}

The base of the robot used within this work is the same as~\cite{us}, i.e. an omnidirectional three-wheeled robot. We can identify a distance $D$ between the center of the robot and each wheel. The radius of the wheel is depicted with $r$, and the angular position of the wheel can be defined with $\alpha_i = i \cdot \frac{2}{3}\pi$ with $i \in [0,1,2]$. With $\mathbf{x}_G = [x_G ~~ y_G ~~ \theta_G]^{\top}$ and $\mathbf{\dot{x}}_G$, we indicate the robot's position and velocity in the global frame. With this information, the kinematic model can be obtained.
\
$\mathbf{T}_R$ represent the rotation matrix between the global and the local frame of the robot:
\begin{small}
\begin{equation}
    \mathbf{T}_R = \begin{bmatrix}
 \cos(\theta_G) & \sin(\theta_G) & 0 \\ 
 -\sin(\theta_G) & \cos(\theta_G) & 0 \\ 
 0 & 0 & 1 \\ 
\end{bmatrix}.
\end{equation}
\end{small}

Moreover, $\mathbf{S_L}$ denotes the static transformation from the robot's frame to each wheel:
\begin{small}
\begin{equation}
    \mathbf{S}_L = \begin{bmatrix}
-\sin(\alpha_1) & \cos(\alpha_1) & D \\ 
-\sin(\alpha_2) & \cos(\alpha_2) & D \\ 
-\sin(\alpha_3) & \cos(\alpha_3) & D \\ 
\end{bmatrix}.
\end{equation}
\end{small}

The wheel velocities $\mathbf{\omega} = [\omega_1, \omega_2, \omega_3]$ can be obtained with the following nonlinear relationship: 
\begin{small}
\begin{equation}
    \mathbf{\omega} = \mathbf{G} \cdot \mathbf{S}_L \cdot \mathbf{T}_{R} \cdot \mathbf{\dot{x}}_G,
\label{eq:kin}
\end{equation}
\end{small}
where $\mathbf{G}$ is the gear reduction ratio matrix, i.e. a $3x3$ diagonal matrix whose diagonal elements are all $1/r$.

With the use of the independent camera control, we introduce an additional degree of freedom. Then, by being $\dot{\gamma}_G$ the rotational velocity of the camera w.r.t. the robot in the global frame, we can control the system with $\mathbf{\dot{x}}_G^{'} = [\dot{x}_G, \dot{y}_G, \dot{\theta}_G, \dot{\gamma}_G]^T$ and modify the just introduced matrices as follows.
\begin{small}
\begin{equation}
    \mathbf{T}^{'}_R=
\begin{bmatrix}
 \multicolumn{4}{r}  0\\ 
 \multicolumn{3}{c}{\mathbf{T}_R} & 0\\ 
 \multicolumn{4}{r} 0\\ 
 0 &  0&  1&1 
\end{bmatrix},
\end{equation}
\end{small}

The last row links together the robot's and camera's rotations. Indeed, when the robot performs a rotation, that movement is transferred to the camera through the joint. 

In both $\mathbf{S}_L$ and $\mathbf{G}$ the differences are minimal.
Indeed, with \begin{small}
\begin{equation}
    \mathbf{S}_L^{'} =
\begin{bmatrix}
 \multicolumn{4}{r}  0\\ 
 \multicolumn{3}{c}{\mathbf{S}_L} & 0\\ 
 \multicolumn{4}{r} 0\\ 
 0 & 0 & 0 & 1 
\end{bmatrix}.
\end{equation}
\end{small}
there is no transformation needed since the camera does not contribute to the robot's base movements. $\mathbf{G}'$ only introduces an element on the diagonal matrix for the corresponding gear ratio.

Thus, we can define our system with
\begin{equation}
    \mathbf{\omega'} = \mathbf{G}^{'} \cdot \mathbf{S}_L^{'} \cdot \mathbf{T}_R^{'} \cdot \mathbf{\dot{x}}_G^{'}
    \label{eq:kin_new}
\end{equation}

\subsection{Uncertainty}
\label{subsec:UNC}

Our system estimates the current state of both camera and robot. We assume that no ground truth data is available at any time. For both estimates we use a sensor fusion method based on EKF~\cite{MooreStouchKeneralizedEkf2014} to obtain position and velocity estimates along with their uncertainty. 
The robot's estimate is obtained using ICP odometry~\cite{Pomerleau12comp}, the two IMUs, and the wheel's odometry with an a-periodic refinement provided by the loop closures events, which is used only for the $x,y$ components for stability reasons.

For the camera's state estimation, we are interested only in the \textit{relative }yaw w.r.t. the robot's base. Note that we cannot use any of the two IMUs directly. Indeed, the IMU attached to the base of the robot will detect \textit{only} the rotational movements of the robot, while the camera one has no way to distinguish what is turning. Therefore, we synchronize the two IMU sources and compute the instantaneous relative measurements. We do not assume any alignment between the two IMUs. Note that, by being both IMUs rigidly attached to the robot's body, any translation offset that is there does not affect the angular velocity measured by them. Indeed, one would only need to account for the relative orientation of the two. The error of these measurements is not taken as the plain sum of variances, which would be rather low, but is fixed at $0.01$. This is because we recognize that perfect synchronization is not usually possible and IMUs can drift and have high spikes in their readings. This generated `differential' IMU is then fused in an EKF estimator with the encoder readings, successfully obtaining an estimate of the relative orientation of the camera with respect to the base of the robot.

The SLAM framework is designed to use \textit{only} the robot's pose and uncertainty within the built graph, while the camera relative pose is taken as a ground-truth. In other words, it is \textit{not} possible to account for \textit{both} the robot and the camera's uncertainties. Nevertheless, we propose a novel solution to this problem. Instead of considering the two as separated entities, what we have done is to merge the two state estimates. In this, the position estimate is exactly the one obtained before, while the orientation is obtained through the sum of both the camera and robot's yaw estimates. The same applies to velocities. This state estimate can be indicated by $\mathbf{x}_{sum} = [x_G, y_G, \psi]$ with the corresponding velocity $\mathbf{\dot{x}}_{sum}$ while the overall variance is computed as the sum of the singular ones and can be directly used within any SLAM framework without any modification. In this way we successfully account also for the uncertainty introduced by the camera's independent movement.

\subsection{Control and constraints}
\label{subsec:CONSTR}

The work in~\cite{us} heavily relies on the assumption that the robot is omnidirectional. We can control such a robot by using~\eqref{eq:kin}. To that end, we utilize an NMPC formulation, generated by the ACADO framework~\cite{Houska2011a}. We keep the system unchanged w.r.t. the one used in~\cite{us} with the same weights and penalties. The NMPC's thorough description is not depicted to avoid the introduction of unnecessary further notation.

At first, we modify the NMPC formulation to account for~\ref{eq:kin_new}. In this, the controlled orientation considered in the framework is the \textit{final global} orientation of the \textit{camera}. The penalty factor associated with the camera's rotation is the same as the one of the robot's rotation. We use this formulation to test both a simultaneous rotation of the camera and the robot and a scenario where only the camera is allowed to rotate. Note that with the latter we are effectively using a semi-holonomic robot that is capable to move only along the $x$ and $y$ axis and, moreover, we linearize the whole system. Finally, to simulate a non-holonomic robot, we impose $v_y = 0$ for the whole duration of the experiment. For that to work we need to split control and enforce a smooth orientation of the robot base to follow the trajectory while the camera reaches the desired orientation.

In all experiments we keep the same maximum rotational speed of the camera, of the robot, and also of the whole system. 
	\section{EXPERIMENTS AND RESULTS}
\label{sec:exp}

\begin{figure}[t]
	\centering
	\includegraphics[angle=0,width=0.3\textwidth]{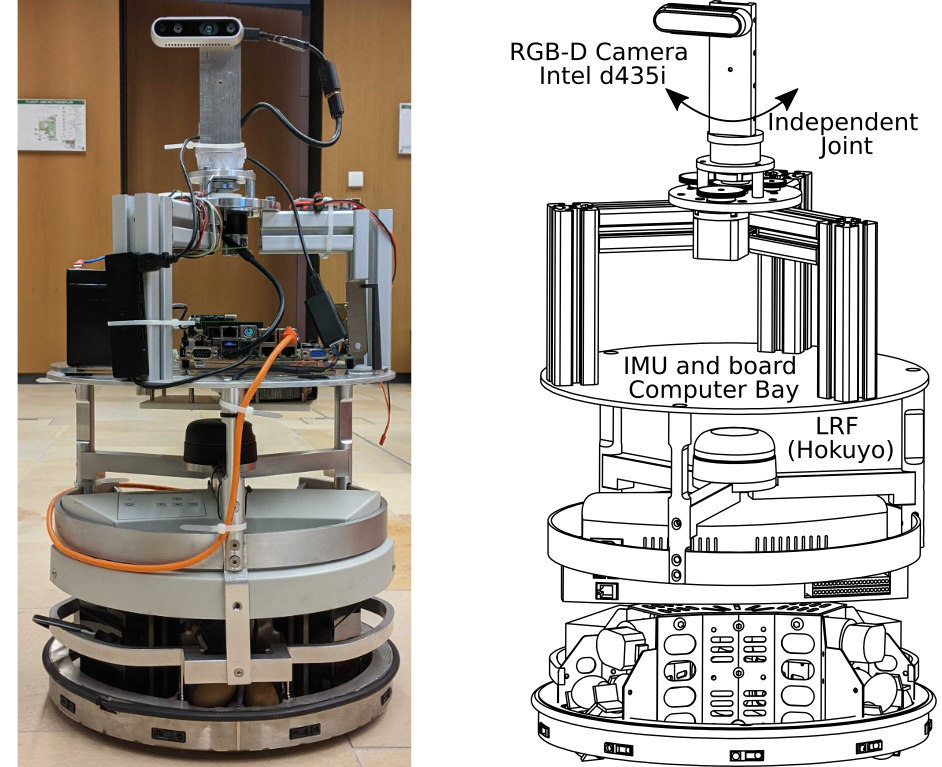}
	\caption{Our robot platform -- Festo Didactic's Robotino with additional structure and hardware. Real robot (left), Gazebo model (right).}
	\label{img:robotschmema}
\end{figure}

Our robot, a Festo Didactic's Robotino 1 (Fig.~\ref{img:robotschmema}), has a 3-wheel omnidirectional drive base. We augmented the robot hardware with an extended physical structure containing a 2D laser range finder (Hokuyo A2M8), an IMU, an RGB-D camera (Intel RealSense D435i) with another embedded IMU unit, and a single-board computer (Intel(R) Core(TM) i7-3612QE CPU @ 2.10GHz). A virtual model of the same robot was created for the simulation experiments (\fig{img:robotschmema} right).

\subsection{Simulation Experiments}

\subsubsection{Setup and Implementation}
\begin{figure}[!ht]
    \centering
    \subfloat[AWS's small house]{
    \includegraphics[scale=.2]{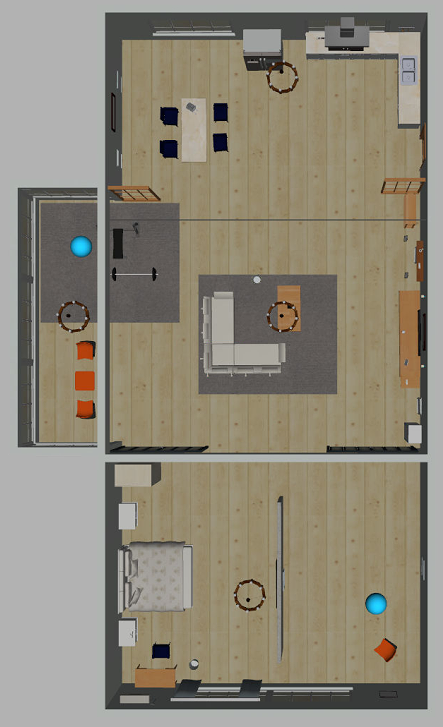}
    \label{fig:aws}
    }
    \quad
    \subfloat[Gazebo's edited Cafè]{\includegraphics[scale=.26]{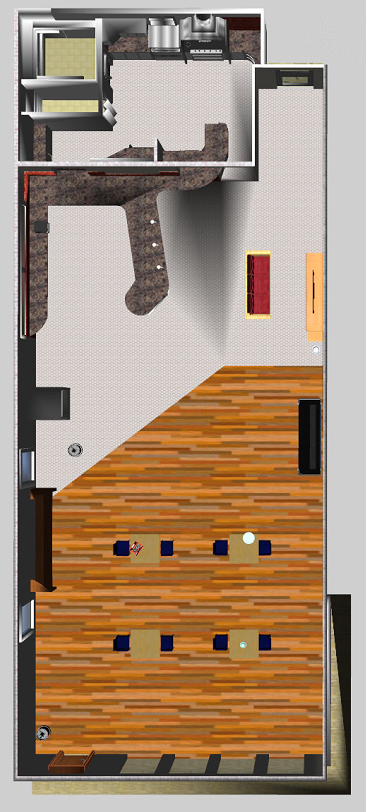}
    \label{fig:gazebo}
    }\caption{Simulated Environments}
    \label{fig:env}
\end{figure}

For validation of the proposed algorithm we use the Gazebo simulator. We test our method in two different environments: i) AWS Robomaker Small House World\footnote{\url{https://github.com/aws-robotics/aws-robomaker-small-house-world}} (Fig.~\ref{fig:aws}), which is $\sim 180 m^2$, and ii) a modified version of the Gazebo's Cafè environment (Fig.~\ref{fig:gazebo}) using 3DGems'~\cite{rasouli2017effect}, which is $\sim 200 m^2$.  

We use real-hardware-like parameters for all the sensors in the simulation. The LRF has an angular resolution of $1^{\circ}$ and provides a complete $360^{\circ}$ sweep. The camera is used at a resolution of $848x480$. Its horizontal FOV is $69.4^{\circ}$, and its maximum sensing depth is $4$ meters. The NMPC horizon is $20$ steps of $0.1$s each. For all experiments the maximum translation speed of the robot is set to $1 m/s$, while the maximum angular speed is $1 rad/s$. The ground truth maps were obtained by using \textit{pgm\_map\_creator}\footnote{\url{https://github.com/hyfan1116/pgm_map_creator}} and edited to remove unwanted artifacts. 
\subsubsection{Metrics}
We use RTAB-Map~\cite{labbe2019rtab} as our V-SLAM back-end, which is a graph-feature-based visual SLAM framework. A 2D occupancy map is generated by the 2D projection of the 3D octomap built through our RGB-D sensor. This allows us to map the obstacles that would be otherwise hidden to the 2D LRF. Maps have a cell size of $0.05 m$ and are compared w.r.t. the ground truth by using several metrics.
\begin{itemize}
    \item The balanced accuracy score (BAC) on the three cells' classes free, occupied and unknown
    \item root mean squared absolute trajectory error (ATE RMSE)
    \item the number of loop closures per meter traveled
    \item the wheels' total overall rotation per meter traveled 
\end{itemize}
With the last point we want to capture the overall energy used by the robot throughout the experiment. Our hypothesis is that, by allowing the independent camera rotation, the system will be capable to balance the movement of both the robot and the independent joint more efficiently. A lower rotational movement of the wheels translates in less energy that has to be used to move the robotic platform. Map entropy is monitored throughout the entire experiment and presented as normalized entropy w.r.t. the actual explored area. All results are presented with mean and standard deviation computed among all the trials.

\subsubsection{Comparisons}
The considered baseline for this work is \textit{A\_1\_2\_3} from~\cite{us}. All approaches use all the three levels of activeness presented in~\cite{us}. 
We compare the baseline, named in this work A\_NC, with our proposed merged state estimate (Subsec.~\ref{subsec:UNC}), which will be called A (without NC) in this work since we are modifying the provided system uncertainty. With this test we can compare if there is any actual difference while we do \textit{not} move the camera. In this way, we are able to obtain a baseline for the state estimate merging strategy. 

Then, we compare our baseline against the different alternatives proposed in  Subsec.~\ref{subsec:CONSTR}. We named these:
\begin{itemize}
    \item HH: \textit{both} the camera and the base can rotate at the same time with the \textit{same} maximum speed
    \item OC: Only the Camera is allowed to rotate. The robot can only perform translational movements. This is a \textit{semi-holonomic} scenario
    \item Y0: the robot is behaving as a \textit{non-holonomic} robot, while the camera is allowed to rotate, i.e. $v_y = 0 m/s$
\end{itemize}
In the first case we are increasing the overall complexity of the system by adding an additional controlled variable. With the other two scenarios we are showing how the active SLAM approach is indeed usable even with other platforms by removing the omnidirectional requirement that was used. 

Finally, we also want to study the effects that our proposed state estimate merging has on those methods. As in the baseline case, the experiments that are suffixed with NC represent the standard approach while the ones without NC use our proposed state estimate.

Simulation results are averaged over 20 different successful trials of ten minutes each for every method with the same starting location. Variability is achieved as a result of different trajectories the robot takes in each one of the runs. Since RTABMap's execution rate is not fixed, but varies based on the movement of the robot, the plots are presented after a post-processing of the data. This consists of a bucketing procedure in time windows of 2 seconds, eventually filled with previous values. All comparisons are recapped in~\tab{tab:comp}.

\begin{table}[!h]
    \centering
    \resizebox{\columnwidth}{!}{
\begin{tabular}{l|c|c|c|c|c|c|c|c}
     & A & A\_NC~\cite{us} & HH & HH\_NC & OC & OC\_NC & Y0 & Y0\_NC \\ \hline
    Omnidirectional & \tick & \tick & \tick & \tick &  &  &  &  \\ \hline
    Camera rotating &  &  & \tick & \tick &  &  &  &  \\ \hline
    Semi-holonomic &  &  &  &  & \tick & \tick &  &  \\ \hline
    Non-holonomic &  &  &  &  &  &  & \tick & \tick \\ \hline
    Merged estimate & \tick &  & \tick &  & \tick &  & \tick & 
    \end{tabular}
}

    \caption{Considered comparisons}
    \label{tab:comp}
\end{table}

\subsubsection{Results}

\begin{figure}[!ht]
    \centering
    \subfloat[Path length evolution]{
    \includegraphics[width=.5\columnwidth]{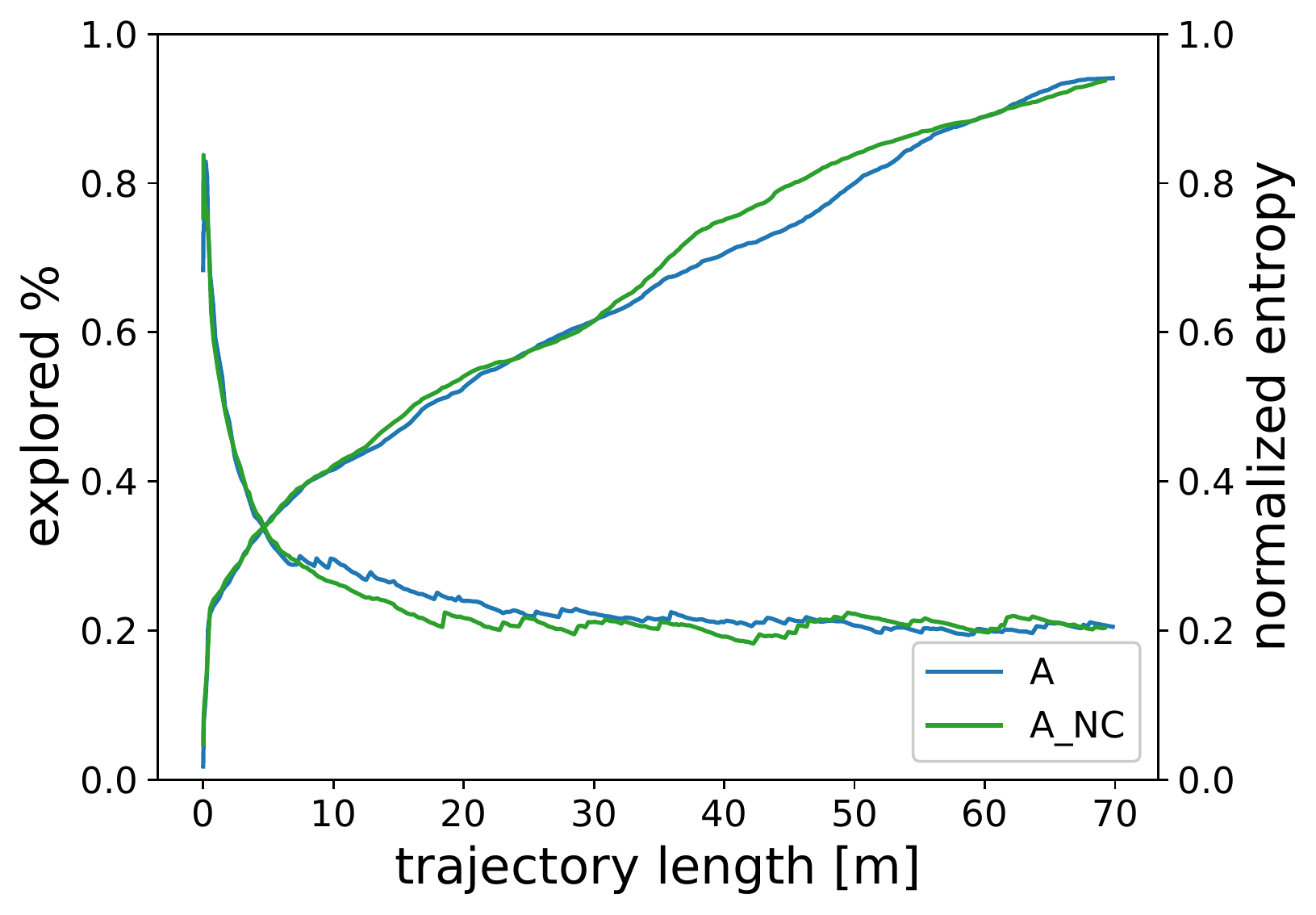}}
    \subfloat[Time evolution]{\includegraphics[width=.5\columnwidth]{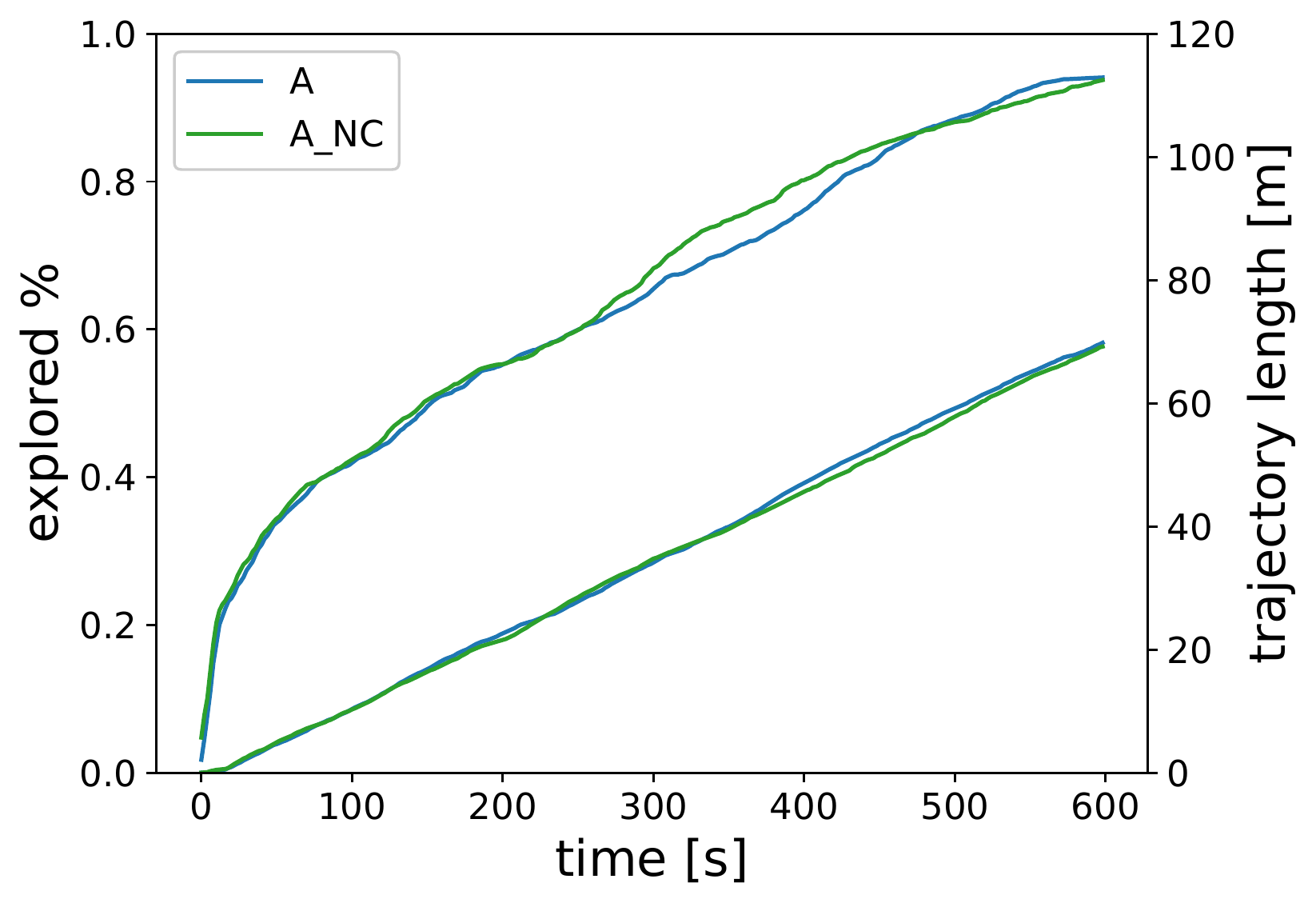}}
    \caption{Cafè: baseline and merged odometry}
    \label{fig:cafe-odom}
\end{figure}

\begin{figure}[!ht]
    \centering
    \subfloat[Path length evolution]{
    \includegraphics[width=.5\columnwidth]{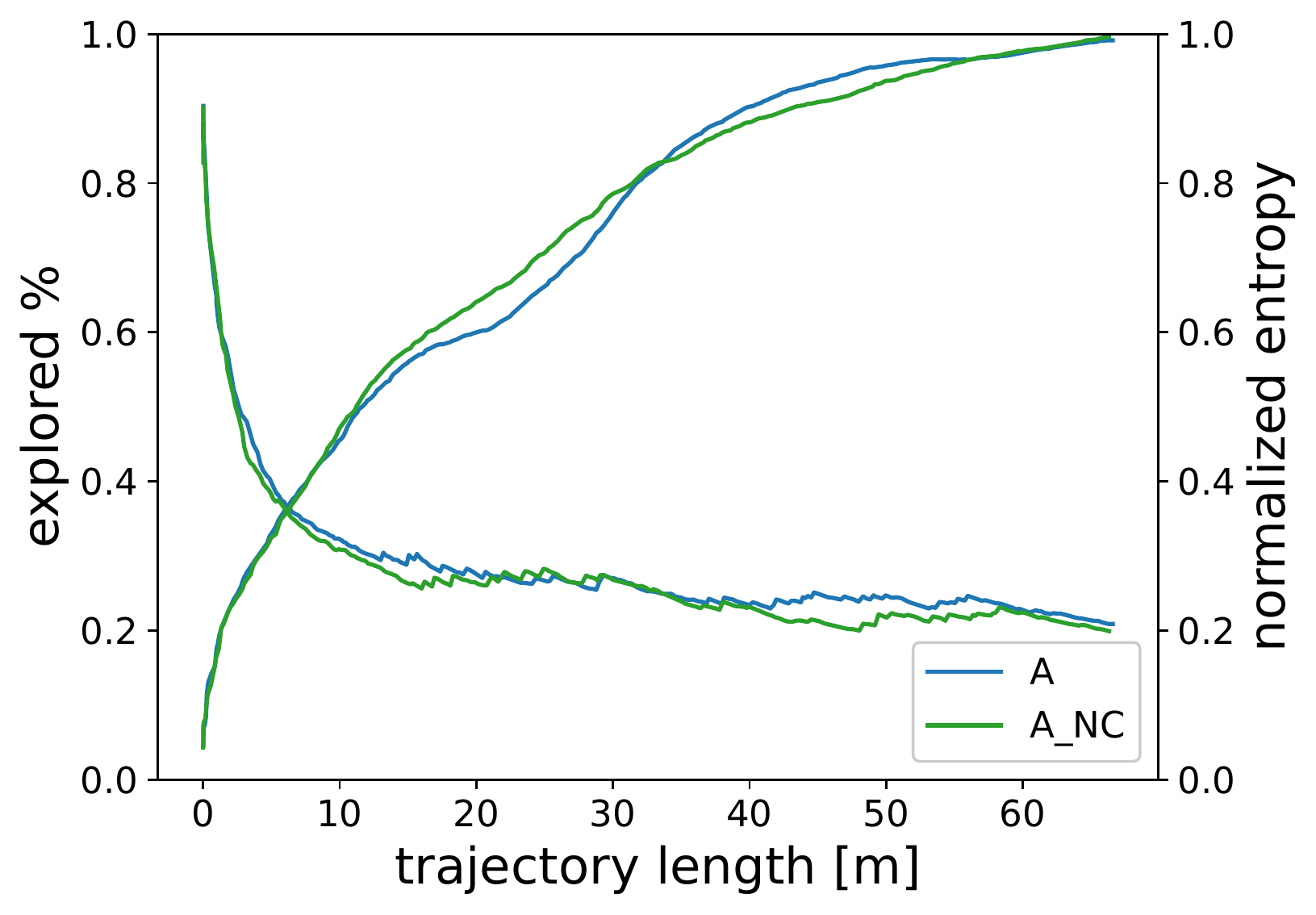}}
    \subfloat[Time evolution]{\includegraphics[width=.5\columnwidth]{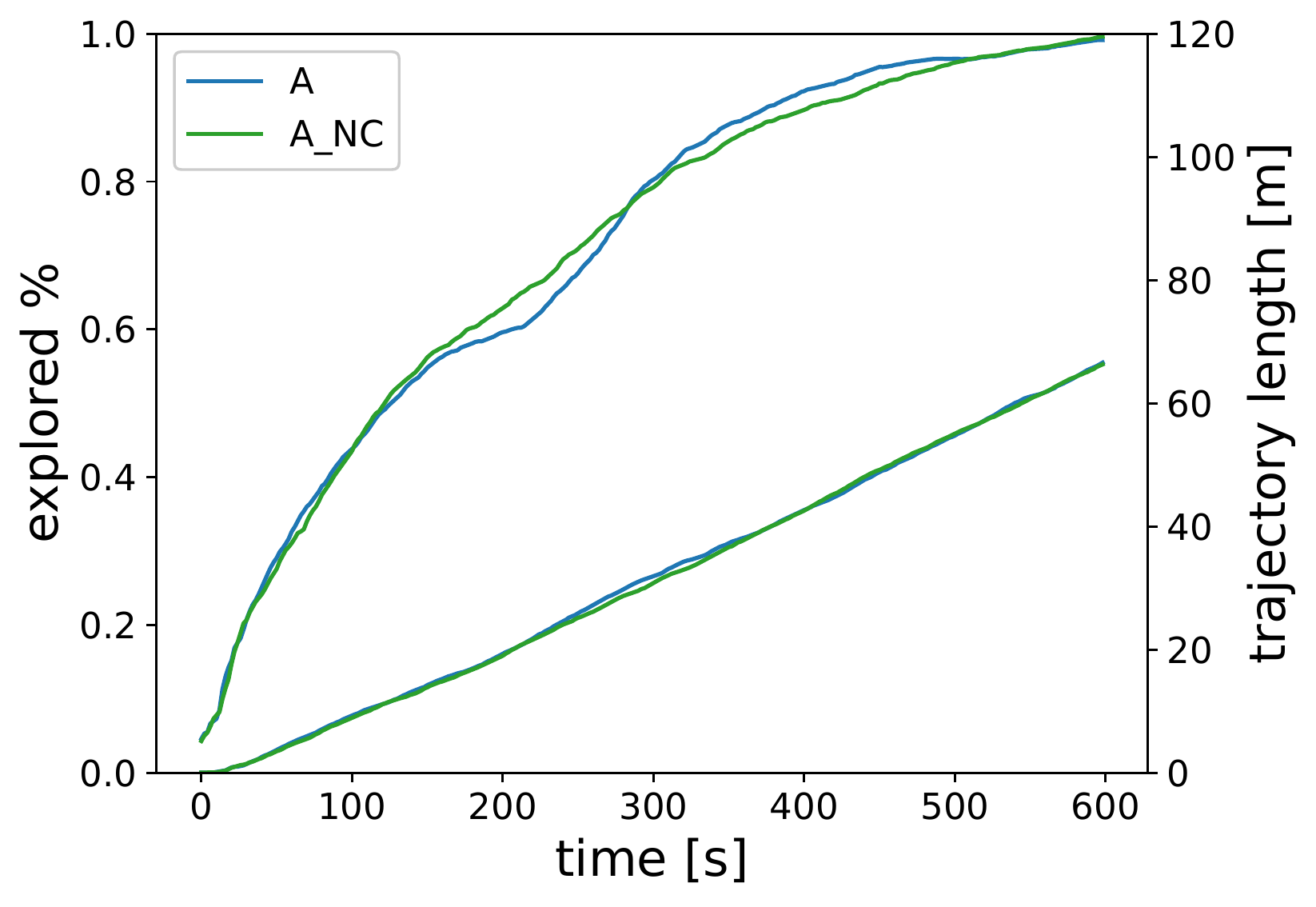}}
    \caption{Small house: baseline and merged odometry}
    \label{fig:sh-odom}
\end{figure}

Considering A and A\_NC, it is clear from the experiments performed on both environments that our proposed way of merging state estimates does not alter the results w.r.t. the considered baseline. There are minimal differences in the mean and standard deviation values (see~\tab{tab:cafe-plat} and~\tab{tab:sh-plat}), but nothing that indicates an overall change of performance. Thus, even though our merged odometry has a higher uncertainty in all the estimates, especially in the yaw velocity, the SLAM framework remains unaffected. As expected, the final entropy, total path length, and exploration speed are similar among the two considered trials as depicted in~\fig{fig:cafe-odom} and~\fig{fig:sh-odom}. 

\begin{figure}[!ht]
    \centering
    \captionsetup[subfloat]{farskip=2pt,captionskip=1pt}
    \subfloat[Path length evolution]{
    \includegraphics[width=.8\columnwidth]{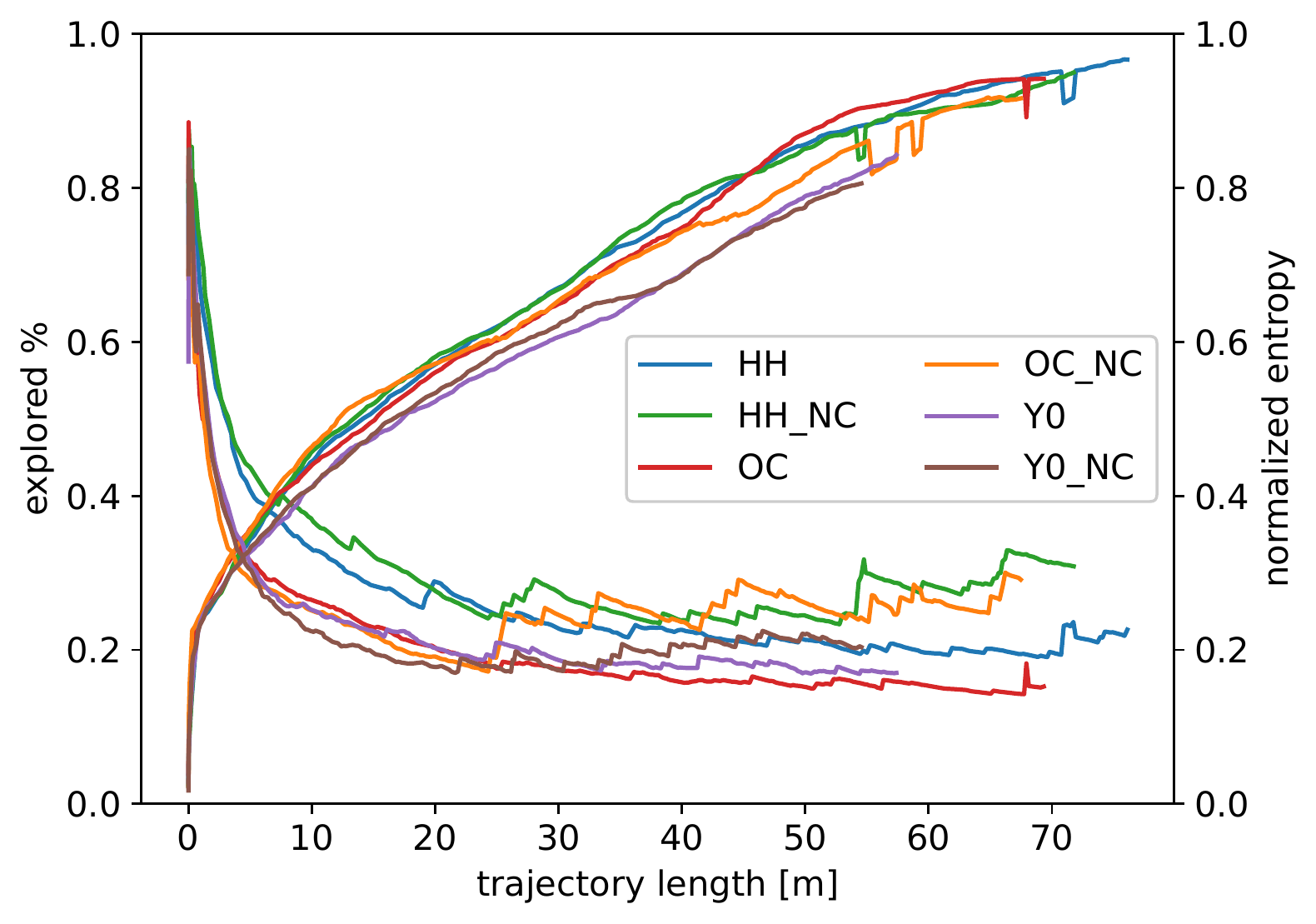}}
    
    \subfloat[Time evolution]{\includegraphics[width=.8\columnwidth]{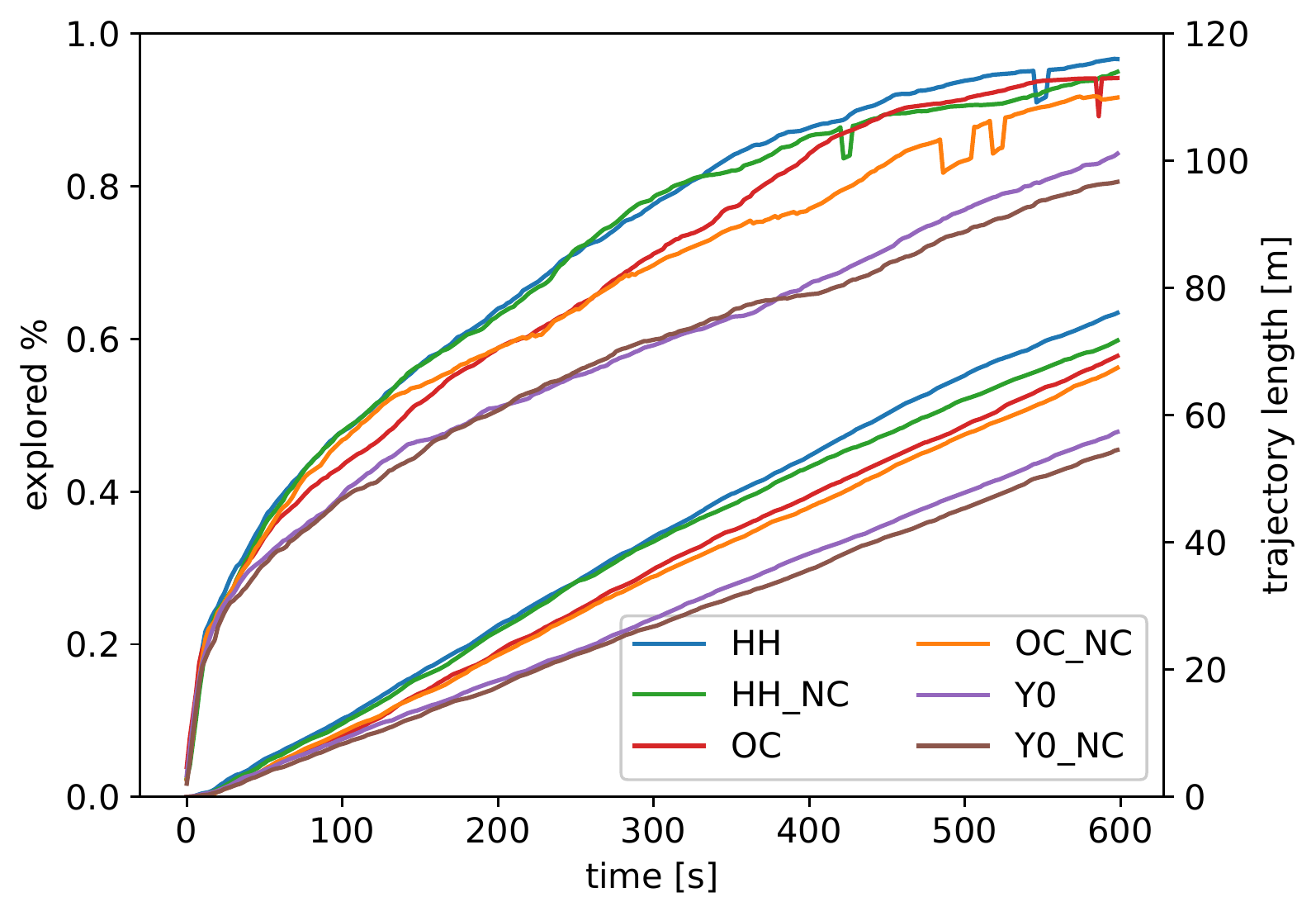}}
    \caption{Cafè: different robotic platforms}
    \label{fig:cafe-plat}
\end{figure}

\begin{figure}[htp]
    \centering
    \captionsetup[subfloat]{farskip=2pt,captionskip=1pt}
    \subfloat[Path length evolution]{
    \includegraphics[clip,width=.8\columnwidth]{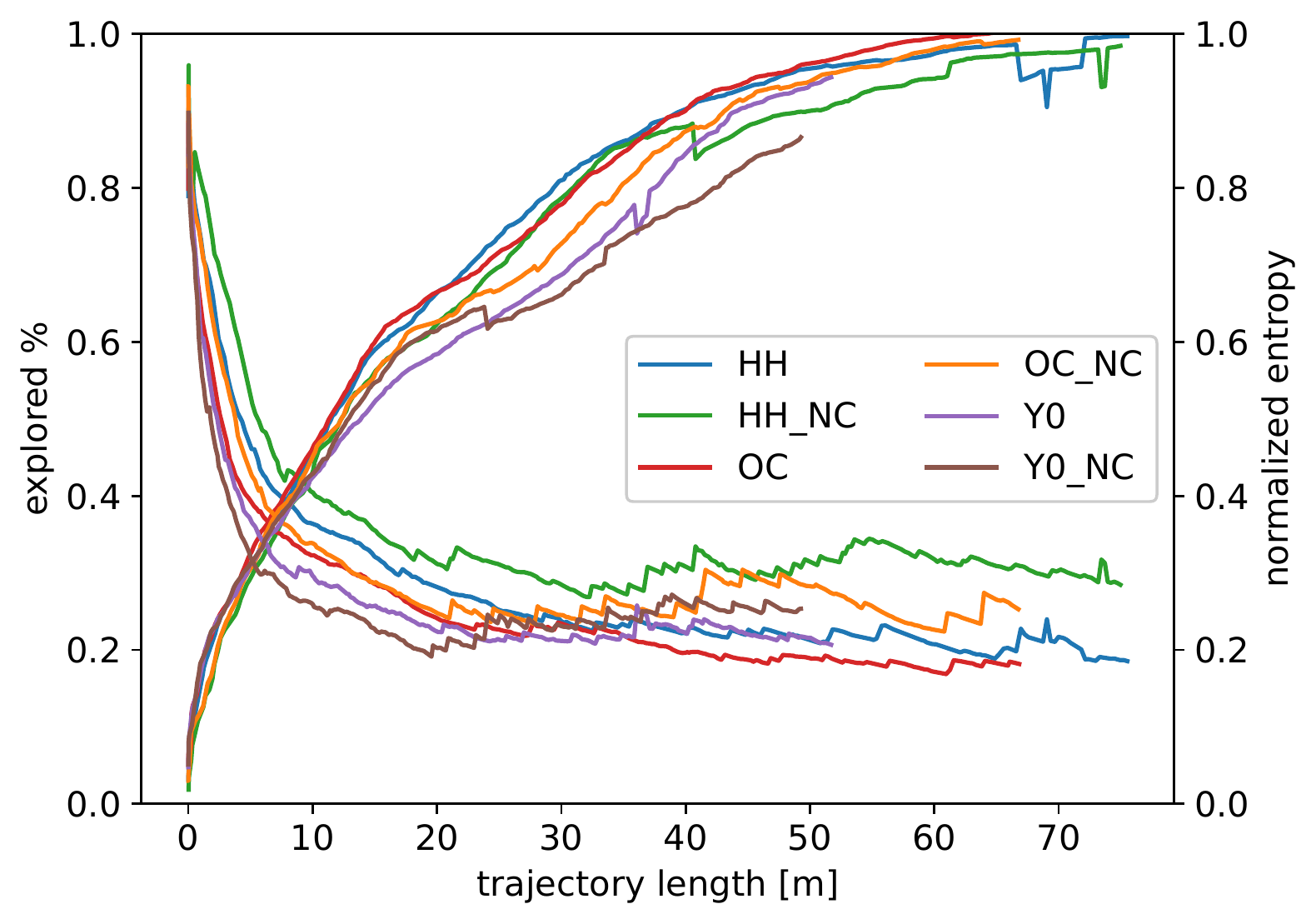}}
    
    \subfloat[Time evolution]{\includegraphics[width=.8\columnwidth]{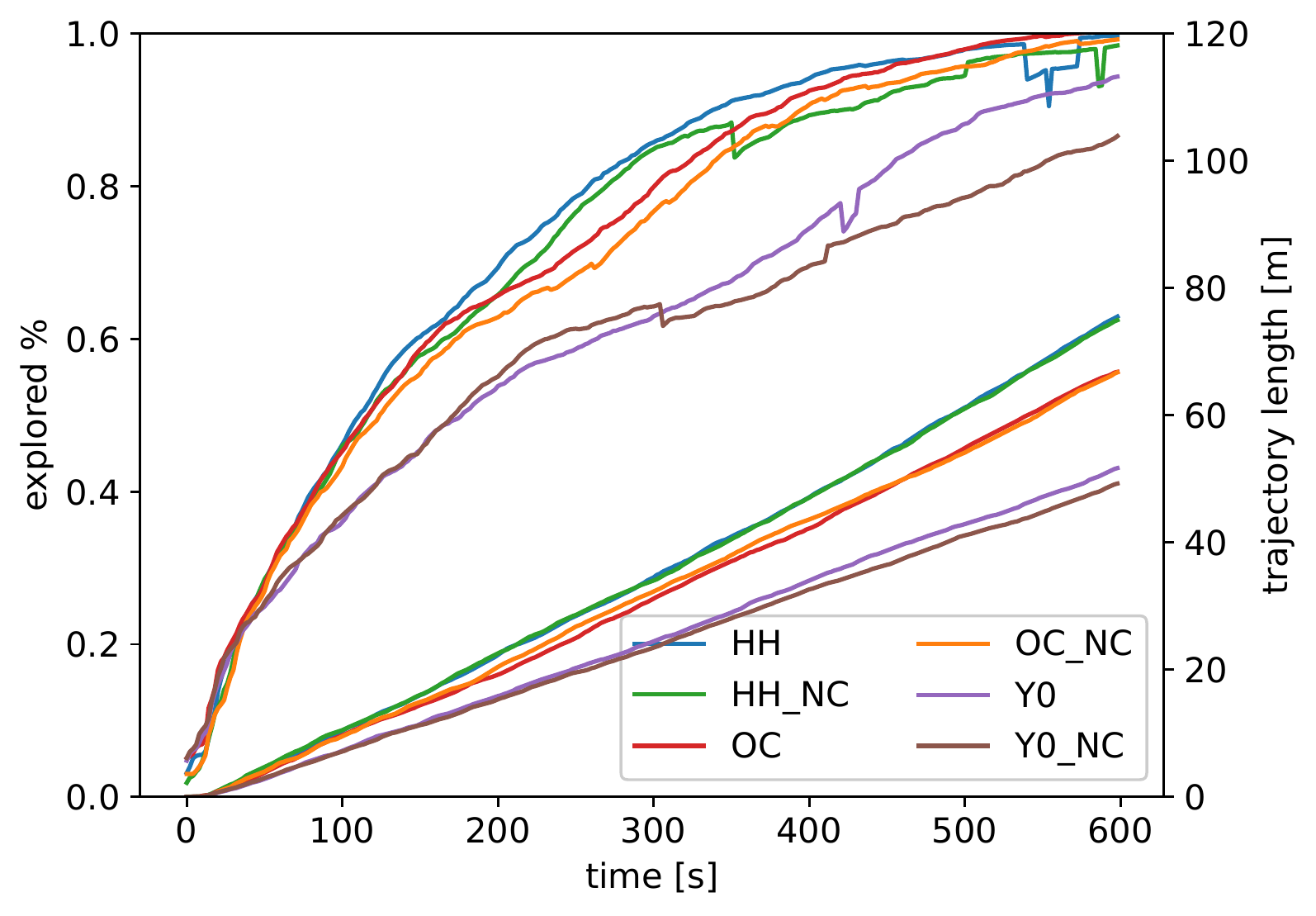}}
    \caption{Small house: different robotic platforms}
    \label{fig:sh-plat}
\end{figure}

When the rotation of the camera is introduced there are some interesting results. First of all, if we compare the two state estimate definitions we can see how, among all experiments, the classical way of considering the camera rotation as perfectly known is worse.
This can be seen in both the environments considered, looking either at the tables~\tab{tab:sh-plat} and~\tab{tab:cafe-plat} or at the figures~\fig{fig:cafe-plat} and~\fig{fig:sh-plat}. In general there is an improvement of the BAC between $\sim1.5 \%$ and $\sim5 \%$ comparing the merged state estimate experiments against the NC counterparts. ATE is also reduced (between $\sim10 \%$ and $\sim40 \%$ lower) and also loop closures are more frequent (between $\sim10 \%$ and $\sim30 \%$ higher) except in the non-holonomic case. 
The greatest difference can be seen once we compare the entropy evolution. In this, when we compare the merged state estimate approach w.r.t. the NC counterparts for all the platforms (HH, OC, and Y0) we can see how the initial evolution is comparable in the first $\sim20 m$ of the trajectory and, after this point, the approaches diverge. The bigger difference can be observed in both the couples HH-HH\_NC and OC-OC\_NC. Finally, Y0 shows the smallest difference in terms of normalized entropy but it has also the shortest traveled path. The trend is divergent with our approach performing better w.r.t. Y0\_NC. Clearly, it is true that the active SLAM method has influence w.r.t. entropy evolution but, since the exploration profiles are equivalent between all the alternative w.r.t. their NC counterparts, we can safely infer that the difference in the performance is not due to this building block. Most likely, the main problem here is the robot registering an orientation for the camera and, when loop closures happen, the drift and the uncertainty of that are not taken into account and have a clear effect on the overall final estimation.

\begin{table}[!ht]
\centering
\setlength\tabcolsep{3pt}
\resizebox{\columnwidth}{!}{
\begin{tabular}{lcccccccc}
\hline
 & \multicolumn{2}{c}{BAC} & \multicolumn{2}{c}{\begin{tabular}{@{}c@{}}Wheels' rotation \\ per traveled meter\end{tabular}} & \multicolumn{2}{c}{ATE RMSE} & \multicolumn{2}{c}{Loops per m} \\
 & \multicolumn{1}{c}{mean} & \multicolumn{1}{c}{std} & \multicolumn{1}{c}{mean {[}rad/m{]}} & \multicolumn{1}{c}{std {[}rad/m{]}} & \multicolumn{1}{c}{mean {[}m{]}} & \multicolumn{1}{c}{std {[}m{]}} & \multicolumn{1}{c}{mean} & \multicolumn{1}{c}{std} \\ \hline
 
A     & $0.774$ & $0.033$ & $66.423$ & $5.934$ & $0.050$ & $0.036$ & $4.868$ & $0.499$ \\
A\_NC & $0.771$ & $0.022$ & $65.781$ & $4.532$ & $0.059$ & $0.028$ & $4.965$ & $0.596$ \\
HH     & $0.793$ & $0.016$ & $57.927$ & $5.198$ & $0.025$ & $0.013$ & $5.180$ & $0.503$ \\
HH\_NC & $0.778$ & $0.027$ & $61.441$ & $8.460$ & $0.042$ & $0.021$ & $4.367$ & $0.508$ \\
OC     & $0.784$ & $0.024$ & $55.494$ & $5.007$ & $0.027$ & $0.034$ & $7.608$ & $0.699$ \\
OC\_NC & $0.755$ & $0.040$ & $56.064$ & $6.023$ & $0.038$ & $0.035$ & $7.081$ & $0.607$ \\
Y0     & $0.728$ & $0.033$ & $55.770$ & $2.941$ & $0.022$ & $0.006$ & $7.452$ & $0.577$ \\
Y0\_NC & $0.705$ & $0.058$ & $57.774$ & $6.904$ & $0.031$ & $0.018$ & $9.486$ & $1.119$
\end{tabular}
}\caption{Cafè environment experiment results}
\label{tab:cafe-plat}
\end{table}

Comparing the three robotic platforms w.r.t. the baseline we notice how, overall, our active SLAM approach presented in~\cite{us} is indeed usable. The first thing to notice is the overall energy reduction that our approach proposed in this work brings. When we rotate both robot and camera, even if we maintain the coupled maximum rotational speed, the system achieves $\sim20 \%$ less total wheel rotation per meter traveled. This implies that the rotation of the camera is capable of reducing the rotation of the robot, further optimizing its movements. This is indeed reflected in every considered metric and in the fact that we perform $\sim10\%$ longer paths within the same time window, suggesting a better optimization and faster reaching of the desired waypoints' orientations. Anyway, this does not result in longer paths since the exploration profiles are comparable.
Moreover, we can see how the semi-holonomic robot (OC) is capable of obtaining even better performance with a lower normalized entropy while keeping a low wheels' rotation amount. In terms of ATE and BAC it is quite similar to the HH method, underperforming slightly in the Cafè environment while getting better results in the Small House. In this case, avoiding the rotation of the base of the robot brings a higher number of loop closures, probably linked to the fact that it is easier for the system to match 2D laser scans. Altogether, these factors indeed help the system in achieving these results. One should also note that, in this scenario, the RTABMap system actually `thinks' that the robot \textit{is} rotating during the experiment, further proving the goodness of both camera rotation (in terms of efficiency) and our merged estimation strategy. It is worth noticing that without the camera rotation, a holonomic robot would have a very inefficient start and stop behaviour to follow the desired heading that continuously changes along the trajectory. 

Finally, the non-holonomic robot (Y0 and Y0\_NC) is the worst in terms of BAC. This is probably linked to the fact that it is capable of exploring a much lower amount of area that lowers the balanced accuracy performance. In this case, we should also note that we did not optimize trajectory or waypoints for a non-holonomic robot to keep the comparison as fair as possible. We believe that performing such optimization would be indeed beneficial, especially in the total amount of explored area. The higher number of loop closures of Y0\_NC can be linked to the slow movement and the lower amount of area seen despite the comparable path length. Anyway, taking into account also the other metrics it seems reasonable to say that Y0 performs better overall.

\begin{table}[!ht]
\centering
\setlength\tabcolsep{3pt}
\resizebox{\columnwidth}{!}{
\begin{tabular}{lcccccccc}
\hline
 & \multicolumn{2}{c}{BAC} & \multicolumn{2}{c}{\begin{tabular}{@{}c@{}}Wheels' rotation \\ per traveled meter\end{tabular}} & \multicolumn{2}{c}{ATE RMSE} & \multicolumn{2}{c}{Loops per m} \\
 & \multicolumn{1}{c}{mean} & \multicolumn{1}{c}{std} & \multicolumn{1}{c}{mean {[}rad/m{]}} & \multicolumn{1}{c}{std {[}rad/m{]}} & \multicolumn{1}{c}{mean {[}m{]}} & \multicolumn{1}{c}{std {[}m{]}} & \multicolumn{1}{c}{mean} & \multicolumn{1}{c}{std} \\ \hline
A     & $0.810$ & $0.019$ & $64.514$ & $5.144$ & $0.056$ & $0.031$ & $4.634$ & $0.525$ \\
A\_NC & $0.818$ & $0.015$ & $63.043$ & $2.540$ & $0.051$ & $0.020$ & $4.692$ & $0.382$ \\
HH & $0.825$ & $0.017$ & $53.635$ & $3.016$ & $0.027$ & $0.013$ & $4.788$ & $0.398$ \\
HH\_NC & $0.808$ & $0.014$ & $54.618$ & $4.872$ & $0.038$ & $0.016$ & $3.659$ & $0.492$ \\
OC & $0.832$ & $0.012$ & $50.232$ & $2.242$ & $0.026$ & $0.017$ & $7.447$ & $0.742$ \\
OC\_NC & $0.811$ & $0.018$ & $51.257$ & $3.002$ & $0.030$ & $0.014$ & $6.597$ & $0.716$ \\
Y0 & $0.785$ & $0.029$ & $56.059$ & $1.861$ & $0.032$ & $0.009$ & $7.334$ & $0.532$ \\
Y0\_NC & $0.742$ & $0.047$ & $61.938$ & $12.097$ & $0.035$ & $0.022$ & $9.642$ & $1.422$ 
\end{tabular}
}
\caption{Small house environment experiment results}
\label{tab:sh-plat}
\end{table}

\subsection{Real Robot Experiments}
The real robot environment has been reconstructed similarly to the one used in~\cite{us}. The experiments consist of five runs of five minutes each. The differences between the results of our baseline (A\_NC) and the results of the same approach presented in~\cite{us} are to be linked to the different environment set-up that caused different paths and to the estimation of the rotation of the camera w.r.t. the base of the robot. Those two main factors combined caused a lower number of loop closures and a higher map entropy. Moreover, since the wheels' encoder readings were not available, we can not directly distinguish between the camera and the robot's rotations. To give an approximation of that we computed the total turn of the base of the robot and of the camera based on the graph nodes. Note that this differentiation is there only in the case the merged estimate is not used. Also, differently from the simulation, this measurement is not fully indicative of the total rotation along the experiment. Therefore, in~\tab{tab:real}, the estimated rotation amounts are expressed separately for all methods in which the merged state estimate is not used (*NC). In the other cases, the \textit{merged} one is being reported in the `Robot's rotation' row. Furthermore, the encoder necessary to read and refine the camera rotational movement through the EKF was not available and the PID controller used caused a considerable delay any time it had to initiate the rotation of the camera. Finally, while the linear speed is the same as the previously run experiments, i.e. 0.3 $m/s$, the angular speed has been increased to 1 $rad/s$ (from 0.5) which caused the robotic platform to perform more harsh and inaccurate rotational movements.

It is shown in~\tab{tab:real} that the total amount of area explored is comparable \textit{but} all the platforms achieve that with shorter paths. Notably, OC\_NC has a much higher exploration amount, $\sim20\%$, w.r.t. the baseline A\_NC. This, combined with the shorter path and the fact that the robot is \textit{not} rotating in OC\*, shows the clear benefits of our approach also in the real world. OC\_NC is less affected by the PID delay because the continuous rotation input from the NMPC causes the camera to almost never stop. The ease of control, the fact that the RTABMap is not aware of the noise related to the camera orientation are to be linked to the outperforming exploration speed of OC\_NC.

Comparing the approximated total estimated rotation, even when the system is forced to not rotate one of the two components (i.e. A\_NC, OC\_NC), it estimates $\sim0.23$ radians per meter traveled of rotation while, ideally, this number should be $\sim0$. Thus, it is clear how the system has in this scenario more difficulties to estimate the map (higher entropy) and recognize places (lower loop closures) w.r.t. the work in~\cite{us}. HH\_NC base's rotation amount is comparable to A\_NC due to the aforementioned PID's delay that had to be compensated by the robot's rotational movement. Despite that, HH\_NC shows some benefits even in these brief experiments by having shorter paths w.r.t. A\_NC but with higher exploration speed. Thus, rotating both the camera and the robot, apart from having a theoretical energy saving advantage and despite the higher control complexity, can bring a benefit to the overall system. We can also notice how the rotational movement per meter is higher in both OC\_NC and HH\_NC  showing a more effective control strategy. 

\begin{figure}[!ht]
    \centering
    \captionsetup[subfloat]{farskip=1pt,captionskip=0pt}
    \subfloat[Path length evolution]{
    \includegraphics[width=.9\columnwidth]{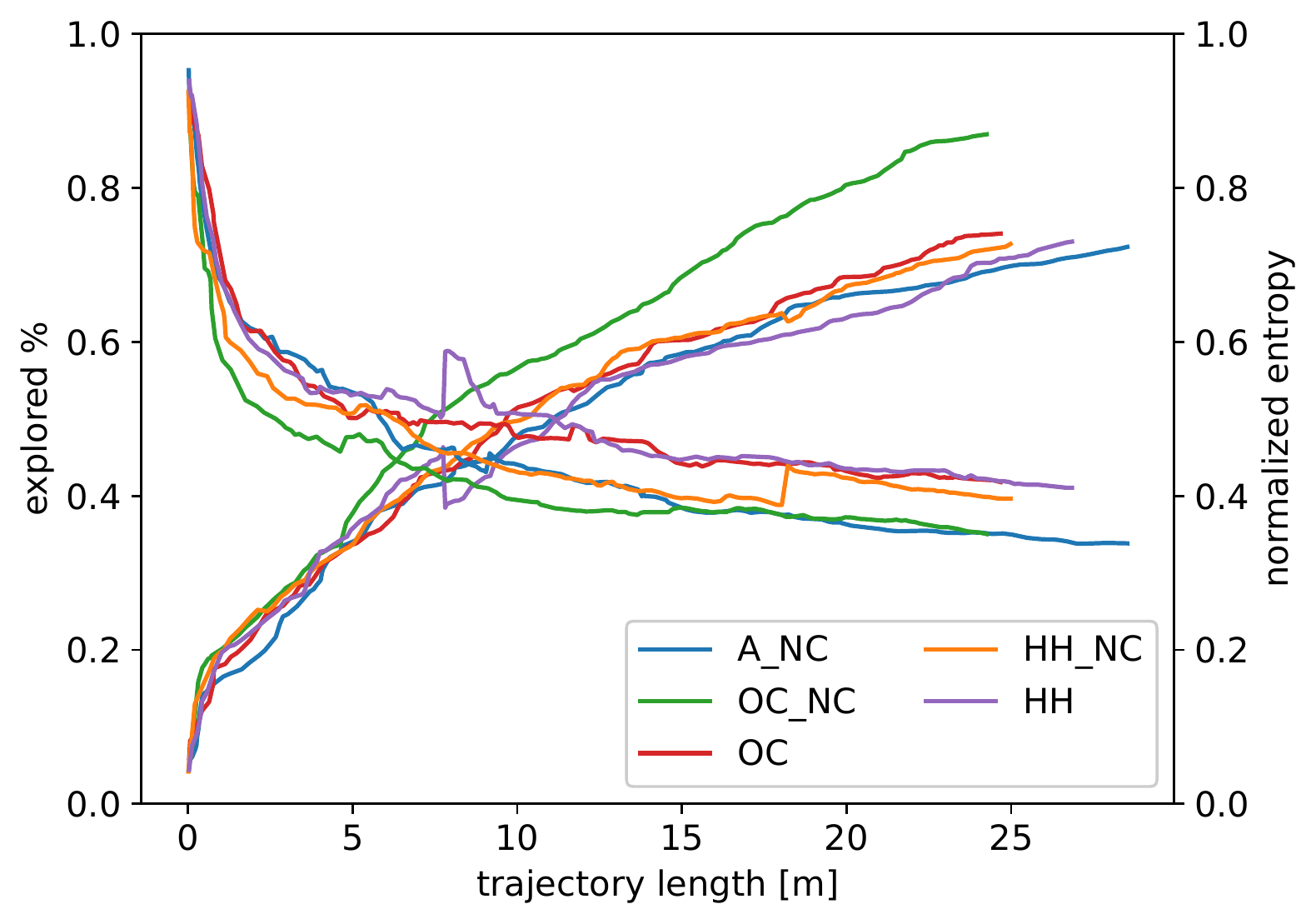}}
    
    \subfloat[Time evolution]{\includegraphics[width=.9\columnwidth]{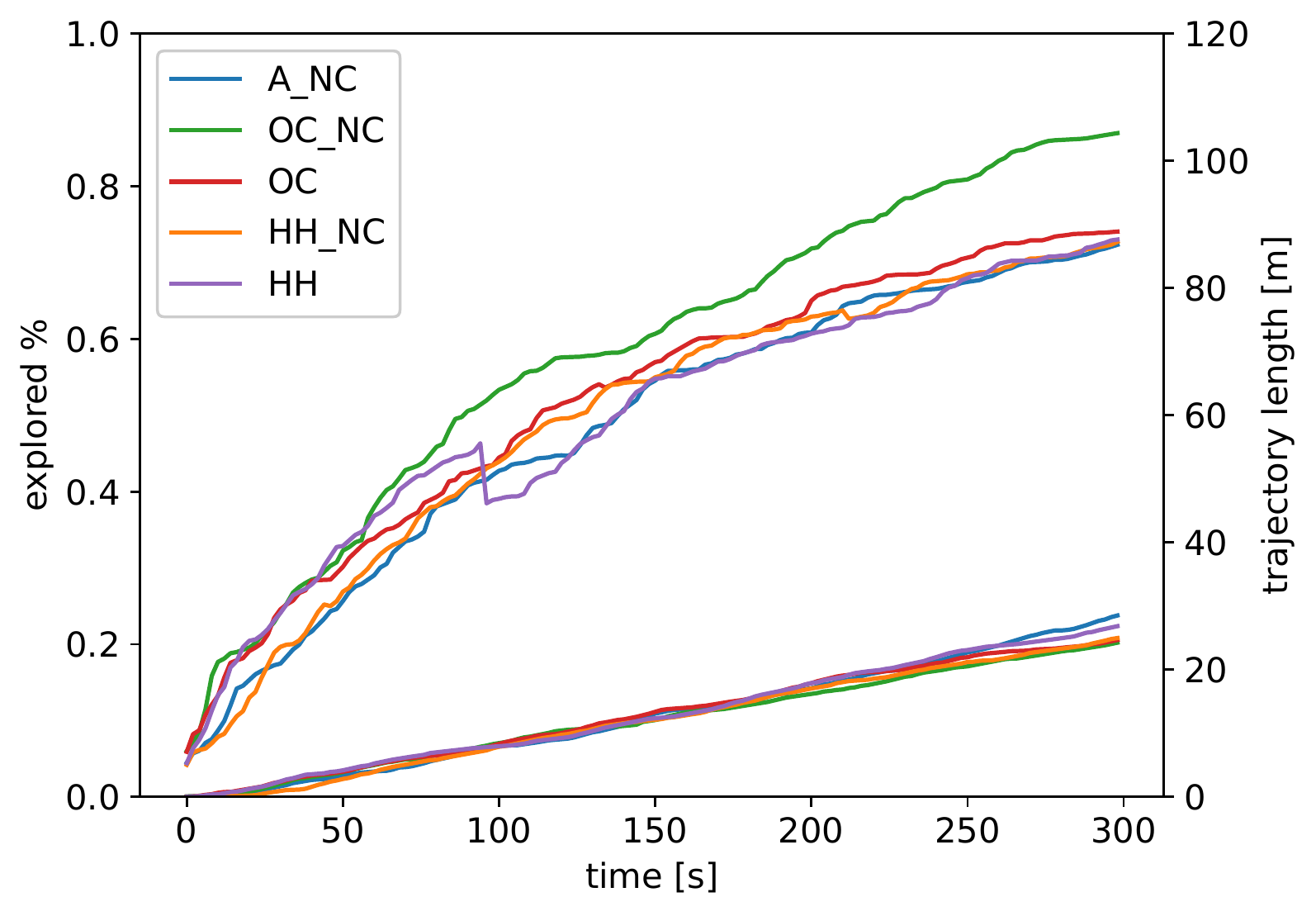}}
    \caption{Real robot experiments}
    \label{fig:real}
\end{figure}

Finally, it is interesting to notice that, despite the missing encoder, the method of merging the state estimates seems promising. We can see from~\tab{tab:real} that the loop closures benefit from this kind of representation, even if the added noise degrades the estimation of the map in terms of normalized entropy by a small amount ($\sim5 \%$, see~\fig{fig:real}). The bigger difference can be seen comparing OC and OC\_NC but, as mentioned before, the performance of OC\_NC could be a bit misleading on its own. 
Overall, given some limitations of our hardware platform, these results show both the usefulness of our approach to different robotic platforms, the benefits of rotating both the camera and the robot simultaneously, and the goodness of the proposed merging strategy for state estimates.

\begin{table}[!ht]
\centering
\setlength\tabcolsep{3pt}
\resizebox{\columnwidth}{!}{
\begin{tabular}{lcccccc}
\hline
 &  & A\_NC & OC\_NC & OC & HH\_NC & HH \\ \hline
\multirow{2}{*}{Area} & mean {[}$m^2${]} & $72.348$ & $86.973$ & $74.072$ & $72.729$ & $73.033$ \\
 & std {[}$m^2${]} & $8.710$ & $7.397$ & $12.473$ & $7.903$ & $12.496$ \\ \hline
\multirow{2}{*}{Robot's rotation} & mean {[}rad/m{]} & $1.675$ & $0.228$ & $3.070$ & $1.684$ & $2.804$ \\
 & std {[}rad/m{]} & $0.250$ & $0.032$ & $0.313$ & $0.362$ & $0.960$ \\  \hline
\multirow{2}{*}{Camera's rotation} & mean {[}rad/m{]} & $0.227$ & $2.580$ & --- & $1.419$ & --- \\
 & std {[}rad/m{]} & $0.023$ & $0.456$ & --- & $0.839$ & --- \\  \hline
\multirow{2}{*}{Path Length} & mean {[}m{]} & $28.526$ & $24.255$ & $24.677$ & $24.978$ & $26.846$ \\
 & std {[}m{]} & $2.516$ & $3.093$ & $2.231$ & $6.794$ & $4.128$ \\  \hline
\multirow{2}{*}{Loops per m} & mean & $0.370$ & $0.920$ & $2.200$ & $0.820$ & $1.570$ \\
 & std & $0.470$ & $0.900$ & $1.530$ & $0.650$ & $2.300$
\end{tabular}
}
\caption{Real robot experiments}
\label{tab:real}
\end{table}

\section{CONCLUSIONS}
\label{sec:conc}

In this work we have shown how we can apply \textit{iRotate} to different robotic platforms, allowing us to use this active SLAM algorithm with a variety of robots and dismissing the omnidirectional requirements. As expected, the map accuracy achieves similar performance w.r.t. the baseline since the exploration strategy remains unchanged and, therefore, the vast majority of cells are mapped equally. Nonetheless, all other metrics show the benefits of our approach in all the considered platforms. Indeed, while the mapping performance is not being affected much by \textit{which} robotic platform is being used it has been shown how an independent camera rotation yields lower energy consumption, comparable or higher amount of loop closures, lower ATE and shorter path lengths. Moreover, the proposed merged state estimate has shown promising results in both simulation and real world experiments despite the limitation of the latter ones. Note that, using a `full' holonomic robot could further optimize the energy consumption in that scenario. A small in-place rotation could indeed align the robot with the movement direction, thus avoiding the rotation movement of one of the wheels. Concerning HH, a diversified control strategy might be applied by seeking energy efficiency with the base and reaching the desired orientation with the camera. This could further reduce the overall energy consumption of the system.
	
%
	

\end{document}